\documentclass[conference]{IEEEtran}
\ifCLASSINFOpdf
\else
\fi
\usepackage{flushend}
\usepackage{balance}
\usepackage{float}
\usepackage{algorithm}
\usepackage{algorithmic}
\usepackage{enumitem}
\usepackage{times}
\usepackage{epsfig}
\usepackage{epstopdf}
\usepackage{amsmath}
\usepackage{amsthm}
\usepackage{amssymb}
\usepackage{amsfonts}
\usepackage{bm}
\usepackage{graphics}
\usepackage{graphicx}
\usepackage{epsfig}
\usepackage{subfigure}
\usepackage{fancyhdr}
\usepackage{array}
\usepackage{multirow}
\usepackage{eucal}
\usepackage{makecell}
\newcommand{\ppttf}{P$^2$T$^2$F}
\newcommand{\comment}[1]{}

\newtheorem{thm}{Theorem}

\begin{document}
%

\title{Simple and Efficient Parallelization for\\ Probabilistic Temporal Tensor Factorization}


\author{Guangxi Li$^1$, Zenglin Xu$^1$, Linnan Wang$^1$, Jinmian Ye$^1$, Irwin King$^2$, Michael Lyu$^2$\\
$^1$ Big Data Research Center, School of Computer Science and Engineering, University of Electronic \\ Science and Technology of China, Chengdu, Sichuan, China; \\ gxli@std.uestc.edu.cn; \{zenglin, wangnan318\}@gmail.com; me@ay27.com\\
$^2$ Department of Computer Science and Technology, The Chinese University of Hong Kong \\Shatian, N.T., Hong Kong;
\{king, lyu\}@cse.cuhk.edu.hk\\
}

\maketitle

\begin{abstract}
\emph{Probabilistic Temporal Tensor Factorization} (PTTF) is an effective algorithm to model the temporal tensor data. It leverages a time constraint to capture the evolving properties of tensor data. Nowadays the exploding dataset demands a large scale PTTF analysis, and a parallel solution is critical to accommodate the trend. Whereas, the parallelization of PTTF still remains unexplored. In this paper, we propose a simple yet efficient \emph{Parallel Probabilistic Temporal Tensor Factorization}, referred to as P$^2$T$^2$F, to provide a scalable PTTF solution.  P$^2$T$^2$F is fundamentally disparate from existing parallel tensor factorizations by considering the probabilistic decomposition and the temporal effects of tensor data. It adopts a new tensor data split strategy to subdivide a large tensor into independent sub-tensors, the computation of which is inherently parallel. We train P$^2$T$^2$F with an efficient algorithm of stochastic Alternating Direction Method of Multipliers, and show that the convergence is guaranteed. Experiments on several real-word tensor datasets demonstrate that P$^2$T$^2$F is a highly effective and efficiently scalable algorithm dedicated for large scale probabilistic temporal tensor analysis.
\end{abstract}


\section{Introduction}
Recent developments of tensor decomposition have great impacts on signal processing \cite{de2007fourth}, computer vision~\cite{Shashua2001Linear}, numerical analysis \cite{beylkin2002numerical}, social network analysis~\cite{tensorXuYQ15}, recommendation systems~\cite{chen2013exact} and etc. A comprehensive overview can be found from the survey paper by ~\cite{kolda2009tensor}.  In particular, automatic recommendation systems significantly benefit from tensor decomposition as it effectively extracts hidden patterns from the multi-way data.

Various tensor decomposition methods have been proposed. The CANDECOMP/PARAFAC decomposition, shorted as CP decomposition, is a direct extension of low-rank matrix decomposition to tensors; and it can be regarded as a special case of Tucker Decomposition by adding a super-diagonal constraint on the core tensor \cite{tucker1966some}. This method, however, fails to consider the fact that the real relational data is evolving over time and exhibits strong temporal patterns, especially in recommendation systems. To resolve this issue, \emph{Probabilistic Temporal Tensor Factorization} (PTTF) \cite{xiong2010temporal}, inspired by probabilistic latent factor models \cite{mnih2007probabilistic,salakhutdinov2008bayesian},
has been proposed by incorporating a time constraint. In contrast with Multi-HDP \cite{porteous2008multi}, Probabilistic Non-negative Tensor Factorization \cite{schmidt2009probabilistic} and Probabilistic Polyadic Factorization \cite{chi2008probabilistic}, PTTF is the only one capturing the temporal effects of tensor data.

\begin{figure}[t]
\centering
 \includegraphics[width=0.4\textwidth,height=4cm]{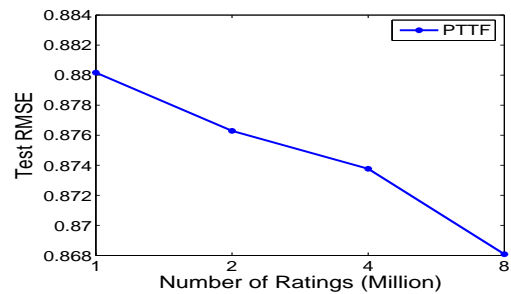}
  \caption{ The RMSE curve for PTTF on MovieLens. The prediction error reduces as the number of ratings grows.} \label{figure_PTTF_scale}
 \end{figure}

The era of big data has also witnessed the explosion of tensor datasets, while the large scale PTTF analysis is important to accommodate the increasing datasets. Figure~\ref{figure_PTTF_scale} demonstrates PTTF achieves better performance as the tensor size increases on the MovieLens data (Table \ref{Table_Summary of real-world datasets}). The result directly sheds light on the necessity of a parallel PTTF solution. Nevertheless, there is a huge gap to be filled.

In this paper,  we present \emph{Parallel Probabilistic Temporal Tensor Factorization} (\ppttf) dedicated for large-scale temporal tensor factorization problems. The core concept of \ppttf{} is to reduce each sequential operation on a large tensor into a set of independent operations on smaller sub-tensors for parallel executions, while still retains the ability to model temporal effects of PTTF. In general, the main contributions of \ppttf{} are as follows:
\begin{itemize}
  \item  \ppttf{} allows parallel solutions to probabilistic tensor decomposition with temporal effects and thus makes PTTF model scalable.
  We demonstrate a new parallelization scheme in \ppttf{} to divide the large-scale problem into several sub-problems for concurrent executions.
  \item In \ppttf, we also design a novel stochastic learning algorithm for parallel ADMM framework to improve the  PTTF model. Specifically, this algorithm calculates the latent feature factors using a substitutive objective function that is convex and can be viewed as an upper bound of the original problem.
  \item The convergence of \ppttf{} is theoretically guaranteed.
\end{itemize}

\section{Related Work}
\comment{Typical collaborative filtering algorithms can be
categorized into two classes: neighborhood methods and factorization methods.
But it is often supplementary between these two class  and the best performance is often obtained by mixing them \cite{bell1996bellkor}. }
Matrix or tensor factorization methods are useful tools in recommendation systems.
One prominent representative factor-based method for recommendation systems is \textit{Probabilistic Matrix Factorization} (PMF) \cite{mnih2007probabilistic}, the latent factors of which can be learned by  maximum likelihood estimation.
Temporal modeling has been greatly ignored in the community of collaborative filtering until the timeSVD++ algorithm is proposed in \cite{koren2010collaborative}. This method demonstrates that the latent features consist of components evolving over time; and such features effectively capture local changes of user preferences. To extend the method to tensors, \cite{xiong2010temporal} proposes the PTTF model to capture the global effect of time shared among users and items.

\comment{
Due to the demand of many large-scale problems, several parallel SGD models have been proposed for PMF problems, such as PPMF \cite{yu2014distributed}, Hogwild \cite{recht2011hogwild} and DSGD \cite{gemulla2011large}.
However, compared to large-scale matrix factorization
\shortcite{wang2016blasx}, the works devoted to large-scale tensors are very few.
}

The increasing demand of modeling data scalability incubates several parallel models for PMF problems, such as PPMF \cite{yu2014distributed}, Hogwild \cite{recht2011hogwild} and DSGD \cite{gemulla2011large}. However, compared to large-scale matrix factorization \cite{wang2016blasx}, there are fewer works devoted to large-scale tensors. In general, the existing large-scale tensor  methods can be categorized into two classes. The first one consists in exploiting sparseness of tensors. For example, the GigaTensor algorithm in \cite{kang2012gigatensor} and DFacTo method in \cite{choi2014dfacto} intend to minimize the number of floating point operations and to handle the size of intermediate data to avoid the intermediate data explosion problem, respectively. The other class of methods consists in distributing the computation load to a number of workers \cite{liavas2015parallel,phan2011parafac,Shang2014tensor}. Unfortunately, these algorithms do not concentrate on modeling the temporal effects of tensor data.


 ADMM is an effective framework for accelerating the optimization of tensor factorization~\cite{2010arXiv1010.0789T,Wang:2015:RKG,Shang2014tensor}. However, they do not target for improving the scalability of tensor factorization algorithms; it also neglects to model the temporal effects in dynamic systems. For example, the parallel ADMM algorithm proposed in \cite{Shang2014tensor} computes tensor decomposition mode by mode; and it is hard to be deployed in online training or distributed training. As a response, we propose \ppttf{} that adapts for the online training or the distributed training. It also works in various computing environments, such as multi-core computers or clusters.

\section{Preliminaries}

\subsection{Notations}
A tensor is a multi-dimensional array that generalizes a vectors (1-dimensional tensor) and a matrix (2-dimensional tensor) to higher order. Like rows and columns in a matrix, an N-dimensional tensor has N modes whose lengths are $I_1$ through $I_N$, respectively. By convention, vectors and matrices are denoted by boldface lowercase letters or uppercase letters with a subscript, e.g., $\bm a$ or $A_i$, and boldface capital letters, e.g., $\bm A$, respectively. We denote higher-order  tensors (order three or higher) by boldface Euler script letters, e.g., $\bm{\mathcal X}$. We also denote the entry of a tensor by the symbolic name of tensor with its indices in subscript. For example, the ($i_1$,$i_2$)th entry of $\bm{A}$ is denoted by $a_{i_1i_2}$, and the ($i_1,\ldots,i_N$)th entry of $\bm{\mathcal X}$ is  denoted by $x_{i_1,\ldots,i_N}$.

\subsection{Tensor Decomposition}

There are several ways to define tensor decomposition \cite{kolda2009tensor,shin2014distributed,de2009survey}. Our definition is based on CP (CANDECOMP/PARAFAC) decomposition, which is one of the most popular decomposition methods. Details about CP decomposition can be found in \cite{kolda2009tensor}. For ease of presentation, we only derive our model in the third-order case, but it can be easily generalized to N-way tensor.

Let $\bm{\mathcal X}\in\mathbb R^{I\times J\times K}$ be a third order  tensor with observable entries $\{x_{ijk}|(i,j,k)\in \Omega\}$, we hope to find the factor matrices $\{\bm{A}\in \mathbb R^{I\times R},\bm{B}\in \mathbb R^{J\times R},\bm{C}\in \mathbb R^{K\times R}\}$ by minimizing the following loss function:
 \begin{eqnarray}\label{formula_order3_CP_decomposition}
 &\min\limits_{\bm A,\bm B,\bm C}& \frac{1}{2}\sum_{(i,j,k)\in \Omega}\left(x_{ijk}- <A_i,B_j,C_k>\right)^2\nonumber\\
 &&+\frac{\lambda}{2}\left(\|\bm A\|_F^2+\|\bm B\|_F^2+\|\bm C\|_F^2\right),
\end{eqnarray}
where $<A_i,B_j,C_k>\equiv \sum_{r=1}^Ra_{ir}b_{jr}c_{kr}$ denotes the inner product of three $R$-dimensional vectors,  $A_i$ denotes the $i$th row of $\bm{A}$, so does $B_j$ and $C_k$. If we solve this model with
\emph{Stochastic Gradient Descent} (SGD), a drastic simplification\cite{bottou2010large}, we can get a SGD-based CP decomposition model. The stochastic process depends on the examples randomly drawn at each iteration.

\subsection{Probabilistic Temporal Tensor Factorization}

%
%


\emph{Probabilistic Temporal Tensor Factorization} (PTTF) model can be considered as the extension of PMF model \cite{mnih2007probabilistic} by adding a specially-constrained time dimension. For the third order tensor $\bm{\mathcal X}$ in (\ref{formula_order3_CP_decomposition}), if the third dimension denotes the time corresponding to the factor matrix $\bm{C}$, we assume the following conditional prior for $\bm{C}$ \cite{xiong2010temporal}:
\begin{eqnarray*}
&C_k\sim \mathcal{N}(C_{k-1},\sigma^2_C\bm{I}_R), \qquad k=1,\ldots,K.&
\end{eqnarray*}
For the initial time feature vector $C_0$, we assume
\begin{eqnarray*}
&C_0\sim \mathcal{N}(\bm \mu_C,\sigma^2_0\bm{I}_R),&
\end{eqnarray*}
where $\bm\mu_C$ denotes a 1-by-$R$ row vector and $\bm{I}_R$ denotes a $R$-by-$R$ identity matrix.
The  PTTF model can be expressed to minimizing the following regularized sum of squared errors (the proof can be seen in \cite{xiong2010temporal}):
 \begin{align}\label{formula_PTTF_orgin}
 &\frac{1}{2}\sum_{(i,j,k)\in \Omega}\left(x_{ijk}- <A_i,B_j,C_k>\right)^2+\frac{\lambda_0}{2}\|C_0-\mu_C\|_2^2\nonumber\\
 &+\frac{\lambda_A}{2}\|\bm A\|_F^2+\frac{\lambda_B}{2}\|\bm B\|_F^2+ \frac{\lambda_C}{2}\sum\limits_{k=1}^K\|C_k-C_{k-1}\|_2^2 ,
\end{align}
where $\lambda_A, \lambda_B, \lambda_C, \lambda_0$  denote regularization parameters.

 Obviously, we can also get a  SGD-based PTTF model using SGD method, but it can not be easily parallelized because of the special constraint on the time dimension.
In contrast, ADMM framework has the properties of  flexibility and tractability and, the most important, can be naturally used to design parallel or distributed learning algorithms for large-scale problems. So, in the next section, we will address the minimization problem in (\ref{formula_PTTF_orgin}) in the framework of ADMM.
In recent years, ADMM has occupied more and more attention and wide range of applications such as matrix completion \cite{yu2014distributed,goldfarb2013fast} and compressive sensing \cite{yang2011alternating}, but as far as we know, few works have been proposed to use parallel ADMM framework  for probabilistic temporal tensor decomposition problems.

\section{Parallel Probabilistic Temporal Tensor Factorization}
In this section, we elaborate \emph{Parallel Probabilistic Temporal Tensor Factorization} (\ppttf). First, we introduce a new data split strategy to divide the whole tensor data into several sub-tensors, and we meticulously reduce a large tensor operation to a set of independent operations toward sub-tensors allowing for concurrent executions. We also extend ADMM to handle these sub-tensors in the training.

\subsection{Data Split Strategy}


The costs of tensor decomposition are closely contingent upon tensor sizes, it is natural for us to split a large tensor data into several independent sub-tensors. In general, we divide the tensor $\bm {\mathcal X}$  into $P$ sub-blocks (or sub-tensors) along the mode with the most dimensions (assuming the first mode without loss of the generality). In this case, each sub-block contains $\frac{I}{P}$ horizontal slices (e.g., $\bm{\mathcal X}^p\in \mathbb R^{\frac{I}{P}\times J\times K}$, $p=1,2,\ldots,P$).

Assuming there are $P$ corresponding local factor matrices for each mode denoted as $\bm A^p$, $\bm B^p$ and $\bm C^p$, respectively. Please note that $\bm B^p$ and $\bm C^p$ share the same size with $\bm B$ and $\bm C$. To better utilize the global variable consensus optimization method, we have a global item latent matrix denoted as $\bm {\overline{B}}$ and a global time latent matrix denoted as $\bm{\overline{C}}$. Since the local matrices $\bm B^p$ and $\bm C^p$ are only coupled with $\bm A^p$, it is feasible to independently update $\bm A^p$, $\bm B^p$ and $\bm C^p$ for each process. This split strategy enables the CP decomposition problem to fit in the parallel ADMM framework.

Algorithm \ref{algorithm_Tensor_split} demonstrates the details of proposed tensor data split strategy.
Please note that it is also possible to divide  a 3-order tensor into $P$ sub-blocks simultaneously along two modes, yet the approach is subject to significant complex loss functions and constraints.
\begin{algorithm}[!t]
\caption{Tensor Data Split}
\textbf{Input:} Tensor $\bm{\mathcal X}\in\mathbb R^{I\times J\times K}$,$P$.

\textbf{Output:} $\bm{\mathcal X}^p\in\mathbb R^{\frac{I}{P}\times J\times K}$.
\begin{enumerate}[itemindent=0.8em]
\item initialize $I_{first}=0$;
\item \textbf{for} $p=1,\ldots,P$ \textbf{do}
\item \quad $I_{last}=\lfloor\frac{pI}{P}\rfloor$;
\item \quad $\bm{{\mathcal X}}^p\leftarrow\bm{\mathcal X}_{(I_{first}+1:I_{last}),:,: }$;
\item \quad $I_{first}=I_{last}$;
\item return  $\bm{\mathcal X}^p$.
\end{enumerate}
\label{algorithm_Tensor_split}
\end{algorithm}

\subsection{Parallel Probabilistic Temporal Tensor Factorization Model}
In this split setting, the minimization problem (e.g., PTTF model) in (\ref{formula_PTTF_orgin}) can be reformulated as the following constrained optimization problem:
\begin{align}\label{formula_PPTTF}
\min\limits_{\begin{subarray}{c}
                      \bm{A}^p,\bm{B}^p,\bm{C}^p, \\
                      C^p_0,\bm{\overline{B}},\bm{\overline{C}}
                    \end{subarray}} & \sum\limits_{p=1}^P\big[f\left(\bm{A}^p,\bm{B}^p, \bm{C}^p\right)
 +g\left(\bm{A}^p,\bm{B}^p, \bm{C}^p,C_0^p\right)\big]\nonumber\\
 \mbox{s.t.}\quad & \bm{B}^p-\bm{\overline{B}} =\bm{0},\nonumber\\
& \bm{C}^p-\bm{\overline{C}}=\bm{0};\forall p\in\{1,2,\ldots,P\}\text.
\end{align}
where
\begin{equation*}
 f\left(\bm{A}^p,\bm{B}^p,\bm{C}^p\right) =\frac{1}{2}\sum_{(i,j,k)\in \Omega^p}(x^p_{ijk}- <A_i^p,B^p_j,C^p_k>)^2,
\end{equation*}
\begin{align*}
 g(\bm{A}^p,\bm{B}^p,&\bm{C}^p,C_0^p)= \frac{\lambda_A}{2}\|\bm A^p\|_F^2+\frac{\lambda_B}{2}\|\bm B^p\|_F^2 \\
 &+\frac{\lambda_C}{2}\sum_{k=1}^K\|C_k^p-C^p_{k-1}\|_2^2 +\frac{\lambda_0}{2}\|C_0^p-\mu_C\|_2^2.
\end{align*}
Here, $\Omega^p$ denotes the ($i,j,k$) indices of the values located in process $p$. $\bm{\overline{B}}$ and $\bm{\overline{C}}$ denote the  global factor matrices.  If we want to generalize it to a $N$-way tensor ($N>3$), we add additional constraints.


We transform the constrained optimization problem in (\ref{formula_PPTTF}) to an unconstrained problem with Augmented Lagrangian Method, and yield the following local objective function:
\begin{align}\label{formula_PPTTF_local}
&L^p(\bm{A}^p,\bm{B}^p,\bm{C}^p,C_0^p, \bm{\Theta}_B^p, \bm{\Theta}_C^p,\bm{ \overline{B}}, \bm{\overline{C}})=f\left(\bm{A}^p,\bm{B}^p, \bm{C}^p\right) \nonumber\\
&+g\left(\bm{A}^p,\bm{B}^p, \bm{C}^p,C_0^p\right)
+l(\bm B^p,\bm C^p, \bm{\Theta}_B^p, \bm{\Theta}_C^p,\bm{ \overline{B}}, \bm{\overline{C}}),
\end{align}
where
\begin{align*}
&l(\bm B^p,\bm C^p, \bm{\Theta}_B^p, \bm{\Theta}_C^p,\bm{ \overline{B}}, \bm{\overline{C}})\\
=\hspace{0.5em} &\mathrm{tr}\left([\bm{\Theta}_B^p]^{\top}(\bm{B}^p-\bm{\overline{B}})\right) +(\rho_B/2)\|\bm{B}^p-\bm{ \overline{B}}\|_F^2\\
&+\mathrm{tr}\left([\bm{\Theta}_C^p]^{\top}(\bm{C}^p-\bm{\overline{C}})\right) +(\rho_C/2)\|\bm{C}^p-\bm{ \overline{C}}\|_F^2.
\end{align*}
Here, $\bm \Theta_B^p$ and $\bm \Theta_C^p$ denote the Lagrangian multipliers, $\rho_B$ and $\rho_C$ are the penalty parameters.

The global objective function is then as follows:
\begin{align}\label{formula_PPTTF_global}
&L(\bm {\mathcal A},\bm{\mathcal B},\bm{\mathcal C},\mathcal C_0, \bm{\Theta}_B,\bm{\Theta}_C, \bm{\overline{B}},\bm{\overline{C}})\nonumber\\
=\hspace{0.5em}&\sum_{p=1}^PL^p(\bm{A}^p,\bm{B}^p,\bm{C}^p, C_0^p, \bm{\Theta}_B^p,\bm{\Theta}_C^p,\bm{\overline{B}}, \bm{\overline{C}}),
\end{align}
where, $\bm{\mathcal A}=\{\bm{A}^p\}_{p=1}^P$, $\bm{\mathcal B},\bm{\mathcal C}, \mathcal C_0, \bm{\Theta}_B$ and $\bm{\Theta}_C$ are similarly defined. The ADMM method will solve this problem by repeating the following steps:
\begin{align}\label{formula_PPTTF_iterations_ABC}
\bm{A}^p_{t+1},\bm{B}^p_{t+1},&\bm{C}^p_{t+1},(C^p_0)_{t+1}\leftarrow \mathop\mathrm{argmin}_{\bm{A}^p,\bm{B}^p,\bm{C}^p,C_0^p} F_{local},\\\label{formula_PPTTF_iterations_B_bar}
\bm{\overline{B}}_{t+1},\bm{\overline{C}}_{t+1}&\leftarrow \mathop\mathrm{argmin}_{\bm{\overline{B}}, \bm{\overline{C}}} F_{global}, \\\label{formula_PPTTF_iterations_ThetaB}
(\bm{\Theta}_B^p)_{t+1}&\leftarrow(\bm{\Theta}_B^p)_t+\rho_B(\bm{B}^p_{t+1}-
\bm{\overline{B}}_{t+1}),\\\label{formula_PPTTF_iterations_ThetaC}
(\bm{\Theta}_C^p)_{t+1}&\leftarrow(\bm{\Theta}_C^p)_t+\rho_C(\bm{C}^p_{t+1}-
\bm{\overline{C}}_{t+1}),\\\nonumber
&\qquad\qquad\qquad \qquad\forall p\in\{1,2,\ldots,P\},
\end{align}
where
\begin{align*}
F_{local}=&L^p(\bm{A}^p,\bm{B}^p,\bm{C}^p,C_0^p,(\bm{\Theta}_B^p)_t, (\bm{\Theta}_C^p)_t, \bm{\overline{B}}_t, \bm{\overline{C}}_t),\\
F_{global}=&\sum_{p=1}^Pl(\bm B^p_{t+1},\bm C^p_{t+1}, (\bm{\Theta}_B^p)_{t}, (\bm{\Theta}_C^p)_{t},\bm{ \overline{B}}, \bm{\overline{C}}).
\end{align*}


These update rules suggest that $\bm A^p$, $\bm B^p$, $\bm C^p$, $C_0^p$, $\bm {\Theta}_B^p$ and $\bm{\Theta}_C^p$ can be locally updated in an independent process. In this case, we dissect the whole tensor factorization problem into $P$ independent sub-problems allowing for concurrent executions. Since we solve the problem under the ADMM framework, \ppttf{} can be viewed as a parallel extension of ADMM applied in PTTF problem.

\subsection{Stochastic Learning Algorithm for \ppttf{}}
To learn the parameters in (\ref{formula_PPTTF}), we need to solve the  above  several steps from (\ref{formula_PPTTF_iterations_ABC}) to (\ref{formula_PPTTF_iterations_ThetaC}). If the optimal $\bm{A}^p$,  $\bm{B}^p$ and $\bm{C}^p$ have been obtained, it is easy to calculate $C_0^p$, $\bm{\overline{B}}$ and $\bm{\overline{C}}$.
By setting the partial derivative of $L^p$ w.r.t $C_0^p$ to zero, we acquire the update rule of $C_0^p$:
\begin{equation}\label{formula_update_C_0^p}
 C_0^p\leftarrow \frac{1}{\lambda_0+\lambda_C}\left(\lambda_CC_1^p+\lambda_0\mu_C\right).
\end{equation}
Since $\bm{\overline{B}}$ is a global variable,  we need to take the partial derivative of $L$ in (\ref{formula_PPTTF_global}) w.r.t $\bm{\overline{B}}$, and set the derivative to zero,
then we can get $\bm{\overline{B}}$. If we set $\left(\bm{\Theta}^p_B\right)_0=\bm{0},p=1,\ldots,P$, we can prove that $\sum^P_{p=1}\left(\bm{\Theta}^p_B\right)_t=\bm{0},t=1,2,\ldots$. Then, the update rules for $\bm{ \overline{B}}$ and $\bm{ \overline{C}}$ ($\bm{ \overline{C}}$ is similar to $\bm{ \overline{B}}$) can be concisely written as:
\begin{equation}\label{formula_update_B_bar}
  \bm{\overline{B}}\leftarrow\frac{1}{P}\sum_{p=1}^P\bm{B}^p, \qquad \bm{\overline{C}}\leftarrow\frac{1}{P}\sum_{p=1}^P\bm{C}^p.
\end{equation}

$\bm{\Theta}_B^p$ and $\bm{\Theta}_C^p$  can be directly updated by formulae  (\ref{formula_PPTTF_iterations_ThetaB}) and (\ref{formula_PPTTF_iterations_ThetaC}). Therefore, how to efficiently compute the factor matrices becomes the key learning part. In the following content of this subsection, we will design a stochastic learning algorithm to solve it.

\subsubsection{Batch Learning}
The update step in (\ref{formula_PPTTF_iterations_ABC}) is actually a PTTF problem with $\bm\Theta_B^p$, $\bm\Theta_C^p$, $\overline{\bm B}$ and $\overline{\bm C}$ fixed. Since $\bm A^p$, $\bm B^p$ and $\bm C^p$ are coupled together and the objective function of the PTTF problem is non-convex, it is not easy to get a satisfied solution. We employ a technique similar to that in \cite{Ouyang2013Stochastic} to resolve this issue by constructing a substitutive objective function, where the factor matrices $\bm A^p$, $\bm B^p$ and $\bm C^p$ are decoupled and can be simultaneously calculated. The convexity of the constructed function, in each iteration, enables us to get the analytical solution of $\bm A^p$, $\bm B^p$ and $\bm C^p$ by setting their gradients to zero.

The substitutive objective function is defined as follows:
\begin{align}\label{formula_PPTTF_local_substitute_H}
&H^p(\bm{\mathcal M}^p,(C_0^p)_t, (\bm{\Theta}_B^p)_t, (\bm{\Theta}_C^p)_t ,\bm{ \overline{B}}_t, \bm{\overline{C}}_t, \tau_t|\bm{\mathcal M}^p_t) \nonumber\\
=&h\left(\bm{A}^p,\bm{B}^p,\bm{C}^p,\tau_t|\bm{A}^p_t,\bm{B}^p_t,\bm{C}^p_t \right)+ g\left(\bm{\mathcal M}^p,(C_0^p)_t\right)\nonumber\\
+&l(\bm B^p,\bm C^p, (\bm{\Theta}_B^p)_t, (\bm{\Theta}_C^p)_t,\bm{ \overline{B}}_t, \bm{\overline{C}}_t),
\end{align}
where
\begin{align}\label{formula_PPTTF_local_substitute_h}
&h(\bm{\mathcal M}^p,\tau_t|\bm{\mathcal M}^p_t)
=f(\bm{\mathcal M}^p_t)+\mathrm{tr}[\nabla_{\bm A^p}^{\top}f(\bm{\mathcal M}^p_t)(\bm A^p-\bm A^p_t)]\nonumber\\
+&\mathrm{tr}[\nabla_{\bm B^p}^{\top}f(\bm{\mathcal M}^p_t)(\bm B^p-\bm B^p_t)]
+\mathrm{tr}[\nabla_{\bm C^p}^{\top}f(\bm{\mathcal M}^p_t)(\bm C^p-\bm C^p_t)]\nonumber\\
+&\frac{1}{2\tau_t}(\|\bm A^p-\bm A^p_t\|_F^2+\|\bm B^p-\bm B^p_t\|_F^2 +\|\bm C^p-\bm C^p_t\|_F^2).
\end{align}
Here, $\bm{\mathcal M}^p=\{\bm{A}^p,\bm{B}^p,\bm{C}^p\}$, $\bm{\mathcal M}^p_t=\{\bm{A}^p_t,\bm{B}^p_t,\bm{C}^p_t\}$ are used for simple writing,
$\tau_t$ is a value that will be related to the learning rate and $f(\bm{\mathcal M}^p_t)$ is already defined in (\ref{formula_PPTTF}).

\begin{thm}\label{theorem_H_L}
Let $D=\{\bm{A}^p,\bm{B}^p,\bm{C}^p\bm |\|A_i^p-(A_i^p)_t\|_2^2\leq \delta^2,\|B_j^p-(B_j^p)_t\|_2^2\leq \delta^2, \|C_k^p-(C_k^p)_t\|_2^2\leq \delta^2\}$, $\delta^2>0$.
Then, $\forall \bm{\mathcal M}^p\in D$,
a suitable $\tau_t$ can always be found to make $H^p(\cdot)$ satisfy the following two properties:
\begin{equation*}
\begin{split}
&H^p(\bm{\mathcal M}^p,(C_0^p)_t, (\bm{\Theta}_B^p)_t, (\bm{\Theta}_C^p)_t ,\bm{ \overline{B}}_t, \bm{\overline{C}}_t, \tau_t|\bm{\mathcal M}^p_t) \\
\geq& L^p(\bm{A}^p,\bm{B}^p,\bm{C}^p,(C_0^p)_t,(\bm{\Theta}_B^p)_t, (\bm{\Theta}_C^p)_t, \bm{\overline{B}}_t, \bm{\overline{C}}_t),\\
&H^p(\bm{\mathcal M}^p_t,(C_0^p)_t, (\bm{\Theta}_B^p)_t, (\bm{\Theta}_C^p)_t ,\bm{ \overline{B}}_t, \bm{\overline{C}}_t, \tau_t|\bm{\mathcal M}^p_t) \\
=& L^p(\bm{A}^p_t,\bm{B}^p_t,\bm{C}^p_t,(C_0^p)_t,(\bm{\Theta}_B^p)_t, (\bm{\Theta}_C^p)_t, \bm{\overline{B}}_t, \bm{\overline{C}}_t).
\end{split}
\end{equation*}
\end{thm}
\noindent The proof of Theorem \ref{theorem_H_L}  can be found in  the supplemental material.
From Theorem \ref{theorem_H_L}, we can find that $H^p(\cdot)$ is an upper bound of $L^p(\cdot)$, and $H^p(\cdot)=L^p(\cdot)$ at the point ($\bm{A}^p_t,\bm{B}^p_t,\bm{C}^p_t$). Fortunately, $H^p(\cdot)$ is convex in ($\bm{A}^p,\bm{B}^p, \bm{C}^p$), and $\bm{A}^p,\bm{B}^p, \bm{C}^p$ are decoupled in $H^p(\cdot)$. Hence, we can optimize the easily-solved constructed function $H^p(\cdot)$ instead of $L^p(\cdot)$ in  (\ref{formula_PPTTF_iterations_ABC}) by setting the gradients to zero, the optimal results are computed as follows:
\begin{align}\label{eq_batch_update_A}
\bm A^p\leftarrow &\frac{1}{1+\lambda_A \tau_t}[\bm A^p_t-\tau_t*\nabla_{\bm A^p}f(\bm{\mathcal M}^p_t)],
\end{align}
\begin{align}
\bm B^p\leftarrow &\frac{1}{1/\tau_t+\lambda_B+\rho_B}\big[\bm B^p_t/\tau_t +\rho_B \bm{\overline{B}}_t\nonumber\\
&-(\bm{\Theta}_B^p)_t -\nabla_{\bm B^p}f(\bm{A}^p_t, \bm{B}^p_t,\bm{C}^p_t)\big],
\end{align}
\begin{align}\label{eq_batch_update_C}
\bm C^p\leftarrow &\bm Q^{-1}\big[\bm C^p_t/\tau_t +\rho_C \bm{\overline{C}}_t+\lambda_C \bm S_C\nonumber\\
&-(\bm{\Theta}_C^p)_t -\nabla_{\bm C^p}f(\bm{A}^p_t, \bm{B}^p_t,\bm{C}^p_t)\big],
\end{align}
 where $\bm S_C=\left[
                  \begin{array}{c}
                    (C_0^p)_t \\
                    \bm 0 \\
                  \end{array}
                \right]
 $ is a $K$-by-$R$ matrix and $\bm Q=(1/\tau_t+\rho_C)\bm I_K+\lambda_C\bm S$ is a $K$-by-$K$ matrix. Here, $\bm S$ denotes a coefficient matrix which shows relationships on the time dimension.
 Actually, $\bm S$ is a tridiagonal matrix and can be described  as follows:
 \begin{equation*}
   \bm S=\left[
                \begin{array}{cc}
                    2\bm I_{K-1}& \bm 0\\
                    \bm 0 & 1\\
                  \end{array}
                \right]+
        \left[
                \begin{array}{cc}
                \bm 0  &  -\bm I_{K-1} \\
                    0 & \bm 0\\
                  \end{array}
                \right]+
        \left[
                \begin{array}{cc}
                   \bm 0 & 0\\
                  -\bm I_{K-1}   & \bm 0\\
                  \end{array}
                \right].
 \end{equation*}

 \comment{
    \begin{equation*}
    \bm S=
    \left[
      \begin{array}{m{0.3cm}<{\centering} m{0.3cm}<{\centering} m{0.3cm}<{\centering} m{0.3cm}<{\centering} c m{0.3cm}<{\centering} m{0.3cm}<{\centering} m{0.3cm}<{\centering} m{0.3cm}<{\centering}}
       2 &-1 & 0 &0  &   \cdots     & 0 & 0 & 0 &0 \\
       -1&2  &-1 &0  & \cdots & 0 &  0& 0 &0 \\
       0 &-1 & 2 &-1 &    \cdots    & 0 &  0&  0& 0\\
        \vdots&  \vdots & \vdots  &  \vdots & \ddots & \vdots & \vdots &\vdots  &\vdots \\
      0 &0  & 0 &0  &   \cdots     &-1&2 &-1& 0\\
      0 & 0 & 0 &0  &  \cdots      & 0 &-1&2 &-1 \\
      0 &0  & 0 &0  &   \cdots     & 0 & 0 &-1&1\\
      \end{array}
    \right].
    \end{equation*}
    }
 A batch learning algorithm for the problem in (\ref{formula_PPTTF}) can be got by combining (\ref{formula_PPTTF_iterations_ThetaB})-(\ref{formula_update_B_bar}) and (\ref{eq_batch_update_A})-(\ref{eq_batch_update_C}).
\begin{thm}\label{theorem_batch_converge}
The batch learning algorithm enables \ppttf{} to converge to a local optimum.
\end{thm}
\vspace{-1ex}
\begin{proof}
From Theorem \ref{theorem_H_L}, we can get
\vspace{-1ex}
\begin{align*}
 &L^p(\bm{\mathcal M}^p_{t+1},(C_0^p)_t,(\bm{\Theta}_B^p)_t, (\bm{\Theta}_C^p)_t, \bm{\overline{B}}_t, \bm{\overline{C}}_t)\\
\leq& H^p(\bm{\mathcal M}^p_{t+1},(C_0^p)_t, (\bm{\Theta}_B^p)_t, (\bm{ \Theta}_C^p )_t ,\bm{ \overline{B}}_t, \bm{\overline{C}}_t, \tau_t|\bm{\mathcal M}^p_t) \\
\leq& H^p(\bm{\mathcal M}^p_t,(C_0^p)_t, (\bm{\Theta}_B^p)_t, (\bm{\Theta}_C^p)_t ,\bm{ \overline{B}}_t, \bm{\overline{C}}_t, \tau_t|\bm{\mathcal M}^p_t) \\
= &L^p(\bm{A}^p_t,\bm{B}^p_t,\bm{C}^p_t,(C_0^p)_t,(\bm{\Theta}_B^p)_t, (\bm{\Theta}_C^p)_t, \bm{\overline{B}}_t, \bm{\overline{C}}_t).
\end{align*}
That is to say, the global objective function $L(\cdot)$ in (\ref{formula_PPTTF_global}) will not increase in each iteration. Furthermore, $L(\cdot)$ is non-convex and  has the lower bound $-\frac{\sum^P_{p=1}\|\bm \Theta_B^p\|_F^2}{2\rho_B}- \frac{\sum^P_{p=1}\|\bm \Theta_C^p\|_F^2}{2\rho_C}$. Hence, our batch learning algorithm will converge to a local optimum.
\end{proof}

\subsubsection{Stochastic Learning}
From the batch learning algorithm, we can find that the update rules  for $\bm{\mathcal M}^p$ will need all ratings related to $\bm{\mathcal M}^p$.  If the number of ratings become very large, the batch learning algorithm will  not be efficient. So, we propose a stochastic learning algorithm to further improve the efficiency, and the update rules for $\bm{\mathcal M}^p$  are as follows:
\begin{align}\label{eq_stochastic_update_ABC}
 A^p_i\leftarrow &\frac{(A^p_i)_t+\tau_t \epsilon_{ijk} ((B^p_j)_t\ast (C^p_k)_t)}{1+\lambda_A \tau_t},\nonumber\\
B_j^p\leftarrow& \frac{1}{1/\tau_t+\lambda_B+\rho_B}\big[(B_j^p)_t/\tau_t +\rho_B ({\overline{B}}_j)_t\nonumber\\
&-(({\Theta}_B^p)_j)_t +\epsilon_{ijk} ((A^p_i)_t\ast (C^p_k)_t)\big],\nonumber\\
C_k^p\leftarrow &\frac{1}{1/\tau_t+2\lambda_C+\rho_C}\big[(C_k^p)_t/\tau_t +\rho_C ({\overline{C}}_k)_t\nonumber\\
&+\lambda_C((C^p_{k-1})_t +(C^p_{k+1})_t)\nonumber\\
&-(({\Theta}_C^p)_k)_t +\epsilon_{ijk} ((A^p_i)_t\ast (B^p_j)_t)\big],
\end{align}
where $\epsilon_{ijk}=x_{ijk}^p-<(A_i^p)_t,(B_j^p)_t,(C_k^p)_t>$.  Hence, the stochastic learning algorithm is a variant of the batch learning algorithm.

By combining the tensor data split strategy and the stochastic update rules stated above, we get a stochastic learning algorithm for our \ppttf{} model. The whole procedure of \ppttf{} is briefly listed in Algorithm \ref{algorithm_PPTTF}, where $\bm{A}=[(\bm{A}^{1})^{\top} \hspace{0.5em}(\bm{A}^{2})^{\top} \cdots (\bm{A}^{P})^{\top}]^{\top}$. Note that the convergence criterion is met when the difference between the train RMSEs of two successive iterations less than some threshold, e.g., $10^{-4}$.

\begin{algorithm}[!t]
\caption{Our \ppttf{} model}
\textbf{Input:} Tensor $\bm{\mathcal X}\in\mathbb R^{I\times J\times K}$, Rank $R$, $MaxIter$, $P$.

\textbf{Output:} $\bm{A}\in\mathbb R^{I\times R},\bm{\overline{B}}\in\mathbb R^{J\times R},\bm{\overline{C}}\in\mathbb R^{K\times R}$.
\begin{enumerate}[itemindent=0.8em]\setlength{\itemsep}{-0ex}
\item use Algorithm \ref{algorithm_Tensor_split} to get $\bm{\mathcal X}^p$;
\item initialize $\lambda_A$, $\lambda_B$, $\lambda_C$, $\lambda_0$, $\rho_B$, $\rho_C$, $\bm{A}^p$, $\bm{B}^p$, $\bm{C}^p$, $C^p_0$, $\bm{\mu}_C$;
\item set  $\bm{\Theta}_B^p$ (and $\bm{\Theta}_C^p$)$=\bm{0}$,  \quad $p=1,2,\ldots,P$;
\item calculate $\bm{\overline{B}},\bm{\overline{C}}$ by (\ref{formula_update_B_bar});
\item \textbf{for} $iter=1,2,\ldots,MaxIter$ \textbf{do}
\item \qquad \textbf{for} $p=1,2,\ldots,P$ parallel \textbf{do}
\item \qquad  \qquad update $C_0^p$ by (\ref{formula_update_C_0^p});
\item \qquad  \qquad \textbf{for}  each $x_{ijk}$ in process $p$ \textbf{do}
\item \qquad  \qquad \qquad update $A_i^p,B_j^p,C_k^p$ by
 (\ref{eq_stochastic_update_ABC});
\item \qquad update  $\bm{\overline{B}},\bm{\overline{C}}$ by (\ref{formula_update_B_bar});
\item \qquad \textbf{for} $p=1,2,\ldots,P$ parallel \textbf{do}
\item \qquad \qquad update $\bm{\Theta}_B^p,\bm{\Theta}_C^p$ by (\ref{formula_PPTTF_iterations_ThetaB}) and (\ref{formula_PPTTF_iterations_ThetaC});
\item \qquad \textbf{if} convergence criterion is met \textbf{then}
\item  \qquad \qquad break;
\item  \qquad update $\tau_t$;
\item return  $\bm{A},\bm{\overline{B}},\bm{\overline{C}}$.
\end{enumerate}
\vspace{-0.5ex}
\label{algorithm_PPTTF}
\end{algorithm}

\subsection{Complexity Analysis}
\ppttf{} mainly needs two steps to update all variables once. The first step updates $\bm{A}^p,\bm{B}^p$ and $\bm{C}^p$. For each value $x_{ijk}$, the time complexity of update $A_i^p$, $B_j^p$ and $C_k^p$ is $O(R)$. Because the total number of observed entries in each process is about  $|\Omega|/P$, the time complexity of step one is $O(|\Omega|R/P)$. The second step needs to update a matrix  of  size $J\times R$ and  a matrix of size $K\times R$ in each process, so the time complexity is $O(\max{\{J,K\}}R)$. In total, the time  complexity of \ppttf{} for each iteration can come  down to $O(|\Omega|R/P+\max{\{J,K\}}R)$.

\begin{figure*}[t]
\centering
\subfigure[S1]{
\label{subfig_box1}
\includegraphics[width=0.3\textwidth]{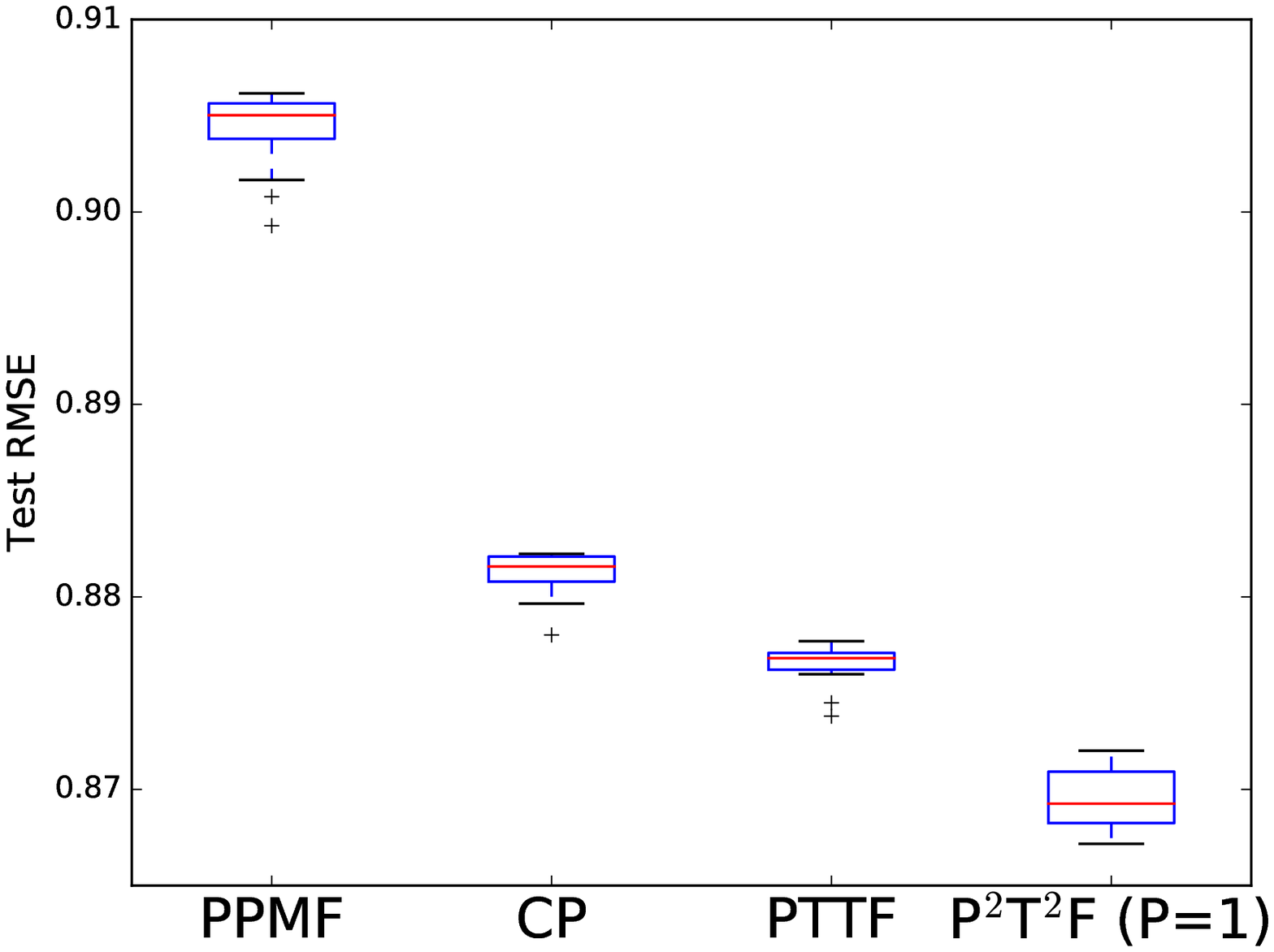}}
\subfigure[S2]{
\label{subfig_box2}
\includegraphics[width=0.3\textwidth]{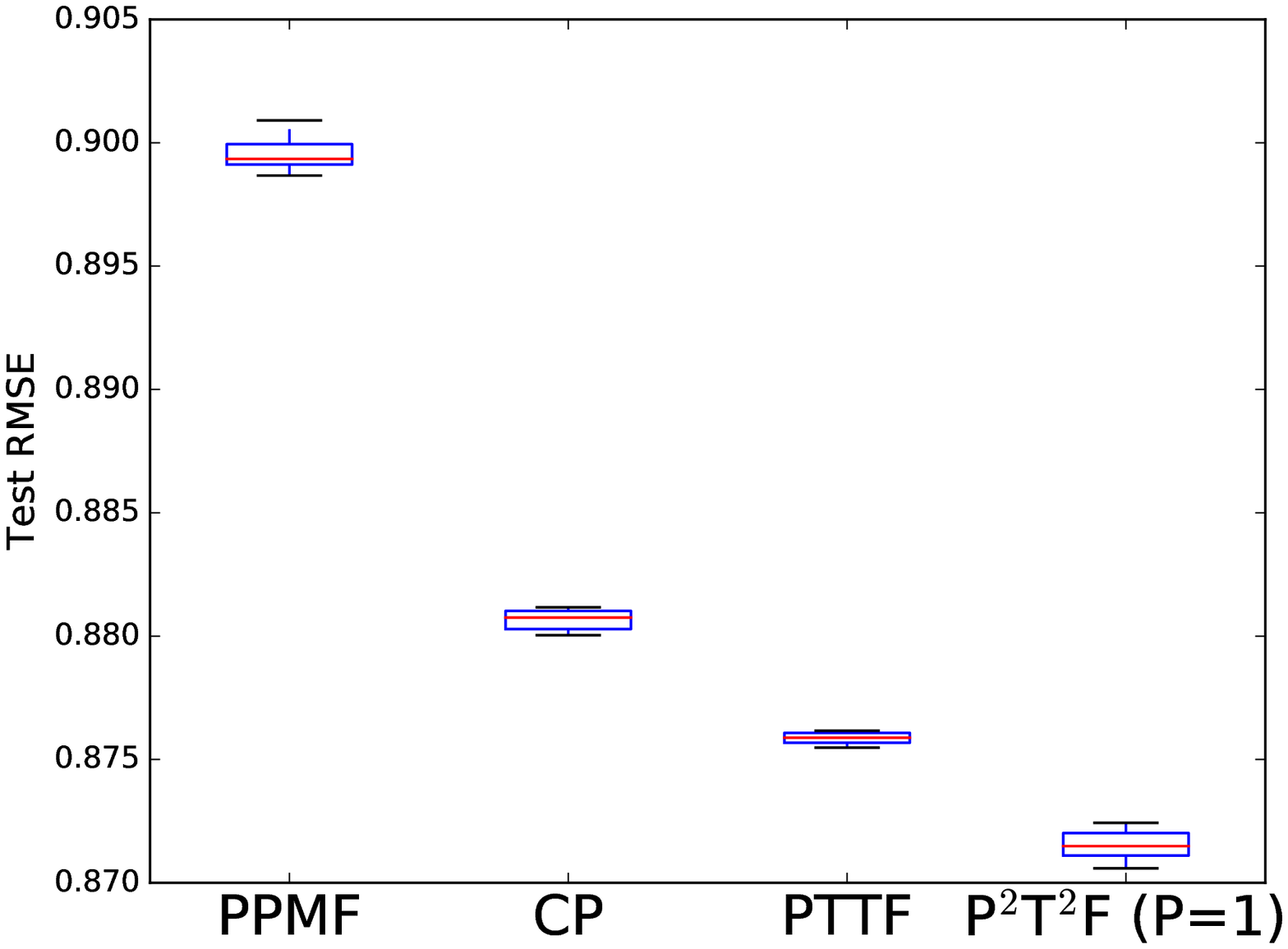}}
\subfigure[S3]{
\label{subfig_box3}
\includegraphics[width=0.3\textwidth]{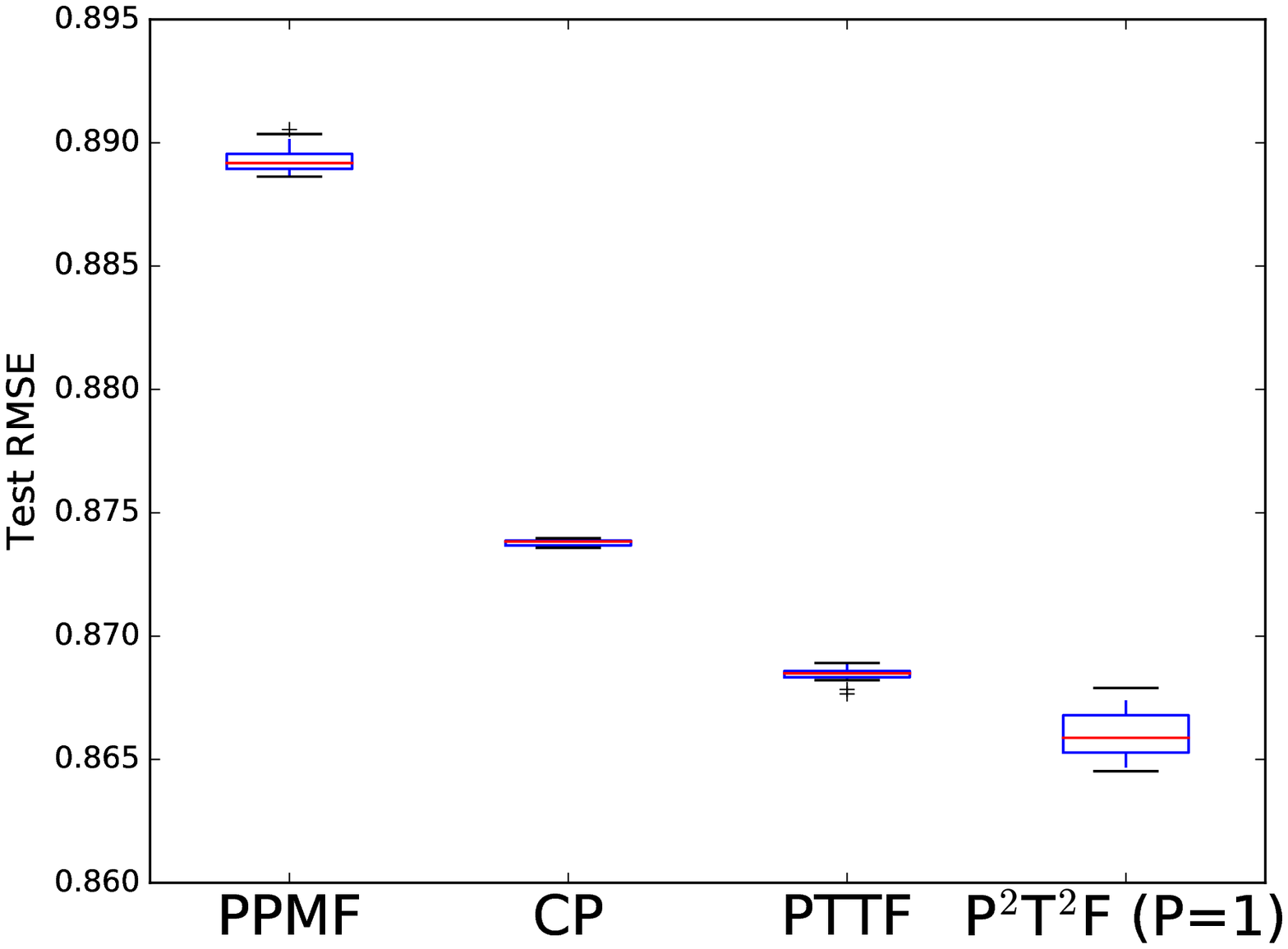}}
 \caption{Box plot of the RMSEs from PPMF, CP, PTTF and \ppttf{} on  three datasets. \ppttf{} can outperform others in one core.} \label{figure_RMSE_box}
\end{figure*}

\begin{figure*}[t]
\centering
\subfigure[S1]{\label{subfig_s1}
 \includegraphics[width=0.3\textwidth]{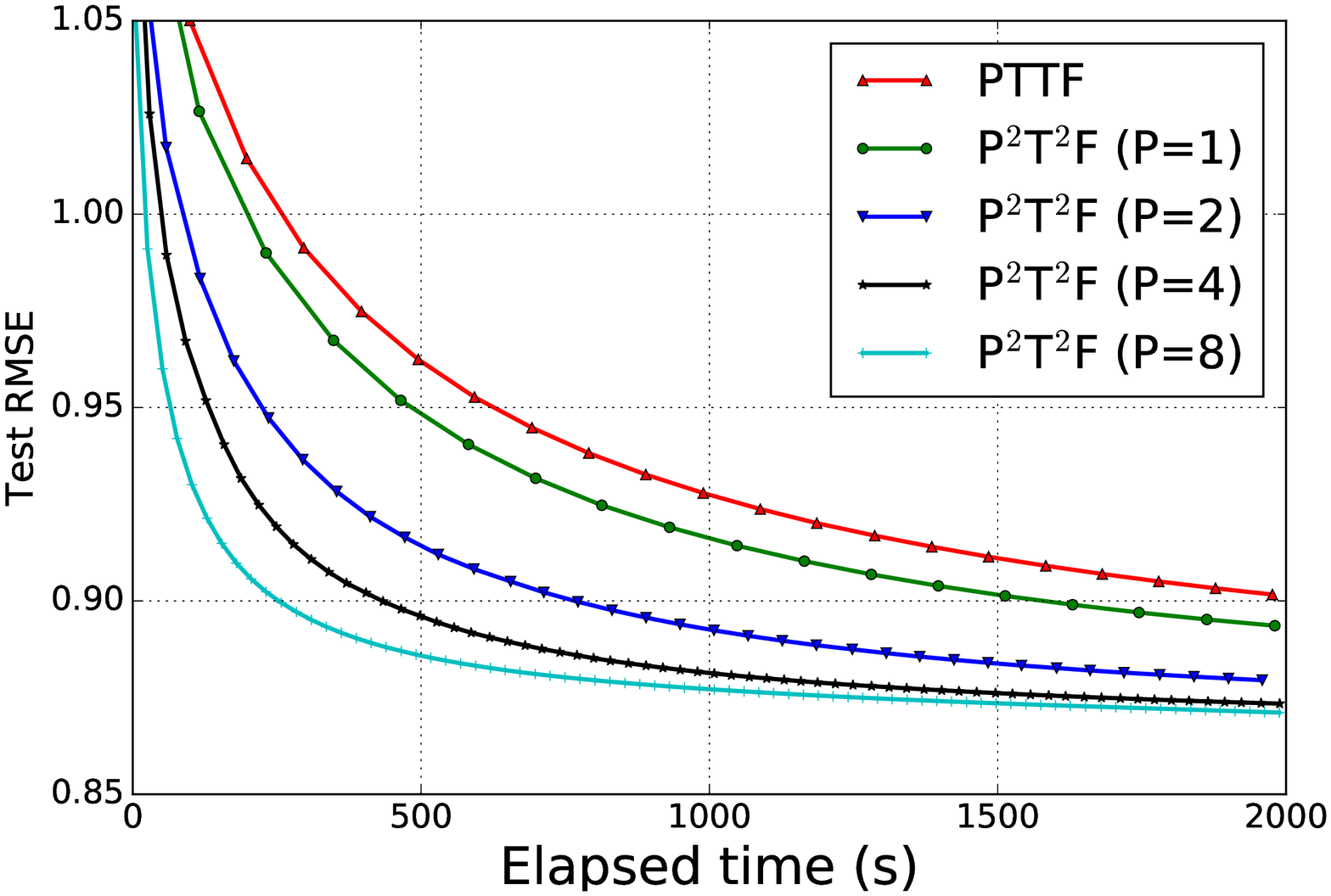}}
\subfigure[S2]{\label{subfig_s2}
 \includegraphics[width=0.3\textwidth]{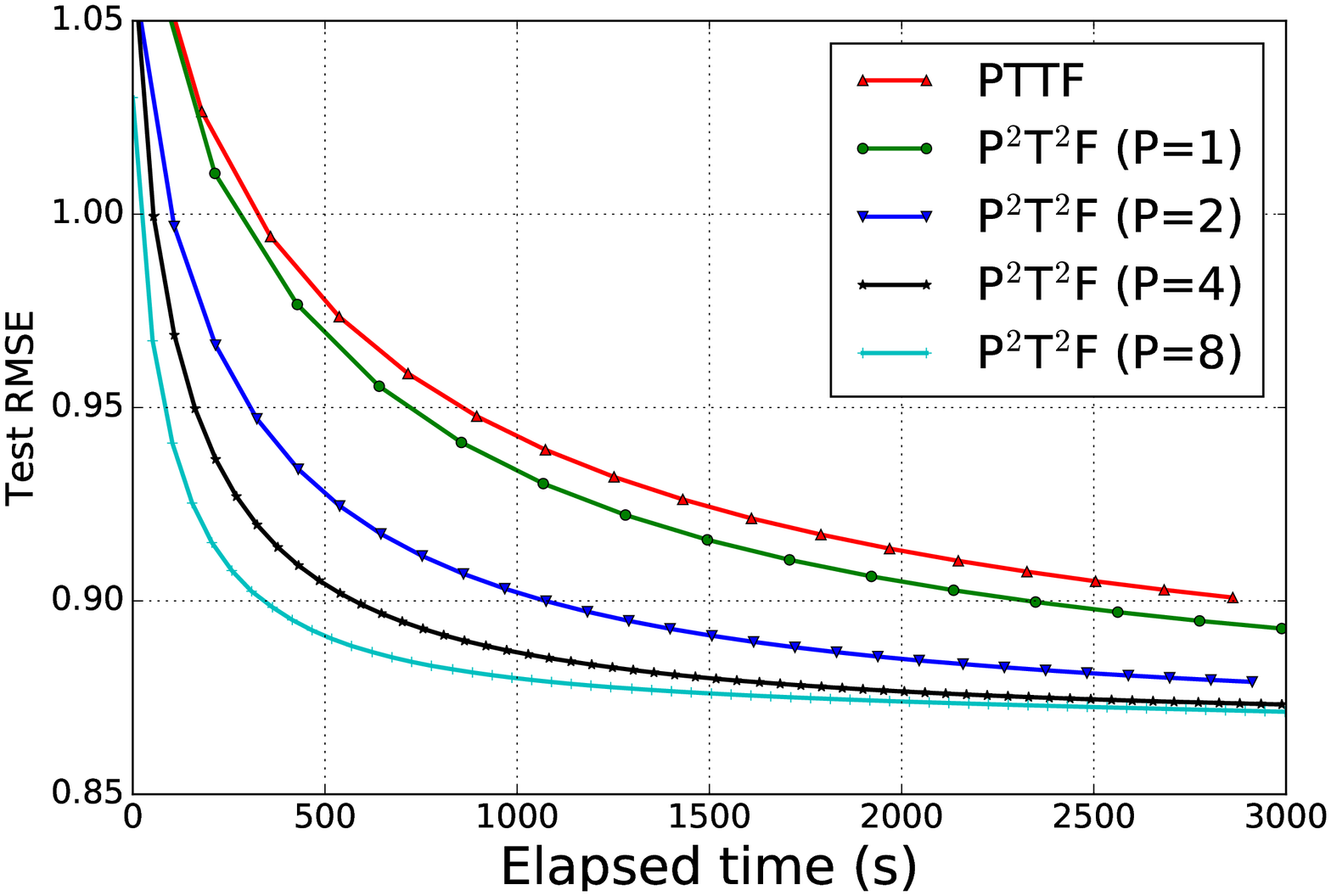}}
\subfigure[S3]{\label{subfig_s3}
 \includegraphics[width=0.3\textwidth]{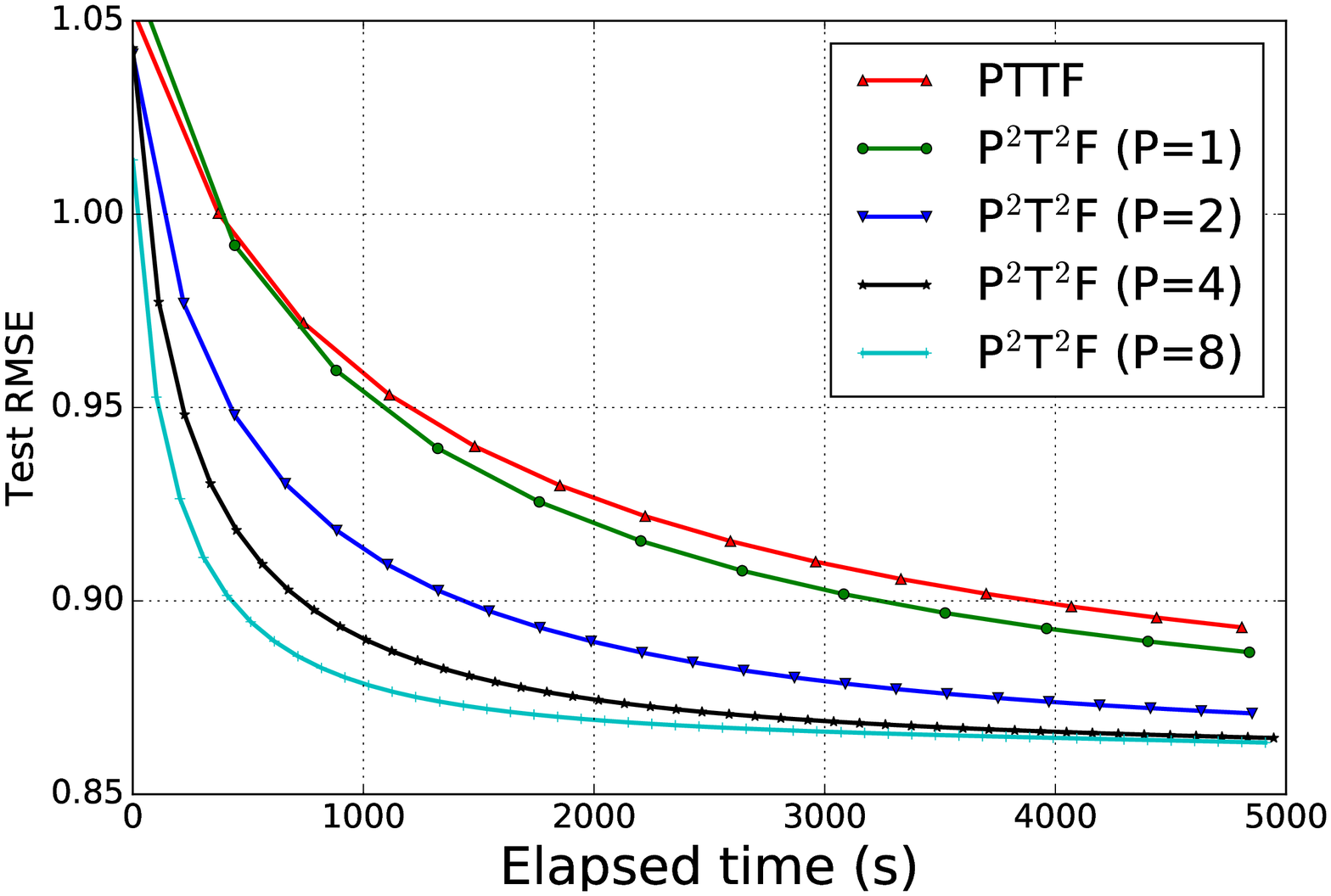}}
  \caption{ The test RMSE curves for PTTF and \ppttf{} with different $P$  on three datasets. } \label{figure_RMSE_Time}
 \end{figure*}

\section{Experimental Results}\label{sec_movie_data}
Our experiments are designed to study the accuracy and efficiency of the \ppttf{} and baselines  on the publicly available real-word datasets. All the experiments are run  on  a 12-core server with 2.60GHz Intel(R) Xeon(R) E5-2630 processor and 64GB of RAM.
\begin{table}[t]
\caption{Summary of real-world datasets}
\renewcommand{\arraystretch}{1.3}
\centering
\begin{tabular}{c|ccc}
   \Xhline{1.2pt}
   & S1 & S2 & S3 \\ \Xhline{1.2pt}
  $I_1$ & 14,012 & 28,060 &56,361\\
  $I_2$ & 19,527 & 24,981 &28,444 \\
  $I_3$ & 242 & 242 &242\\
  $\#Train$ & 1,851,291 & 3,739,047 &7,566,903\\
  $\#Test$ & 205,699 & 415,449 &840,766\\
  \Xhline{1.2pt}
\end{tabular}\label{Table_Summary of real-world datasets}
\end{table}
\paragraph{Datasets and Parameter Settings}
The real-word tensor data used in  our experiments  are public collaborative filtering datasets: Movielens ml-latest \footnote{http://grouplens.org/datasets/movielens/}, which is movie rating data from MovieLens, an online movie recommender service. In order to study \ppttf{}'s parallel performance,  we process  it into three  3-order tensors,  where  each mode  correspond to users, movies and calendar month, respectively,   with the restriction of minimal 20 ratings per user.  The rates range from 0.5 to 5, and the details are summarized in Table \ref{Table_Summary of real-world datasets}. We set $\lambda_A=\lambda_B=\lambda_C=0.01$ and $\rho_B=\rho_C=0.5$. Since it is difficult to compute the exact value for $\tau_t$,
we approximately update it as $\tau_{t+1}=\tau_t*\beta (0 < \beta< 1)$ for
the $t$-th iteration. We also set a threshold $\alpha$. When $\tau_t\leq \alpha$, we stop decreasing $\tau_t$. We set  $\tau_0=0.0005$, $\beta=0.9$ and $\alpha=0.0001$. The hyperparameters are all determined by cross validation. We choose $R=20$ here.

\paragraph{RMSE and Efficiency}
We use the root mean squared error (RMSE) to evaluate our \ppttf{} model and baselines.  We first examine the significance of the improvement of \ppttf{} over the CP, PTTF and PPMF  model on these datasets in one core by repeating the prediction tasks 12 times using different random initializations.
Figure \ref{figure_RMSE_box} shows the resulting box plot of test RMSEs on S1-S3 at the  moment when the convergence criterion of all these methods is satisfied. We can see that  \ppttf{} model outperforms the baselines  in all runs.
\ppttf{} takes advantages of the stochastic learning algorithm, where it handles the latent factors using a surrogate objective function and makes them decoupled and easily computed. Therefore, it reaches a better RMSE than the conventional SGD-based PTTF method.
In particular, the PTTF model outperforms CP model because of the temporal effects in the probabilistic decomposition. The PPMF model only considers the users and movies get the worst result. CP and PTTF model have a smaller degree of dispersion because they have fewer parameters than PPMF and \ppttf{} methods. Therefore, they are less sensitive random initializations.

\begin{figure}[t]
  \centering
  \includegraphics[width=0.28\textwidth]{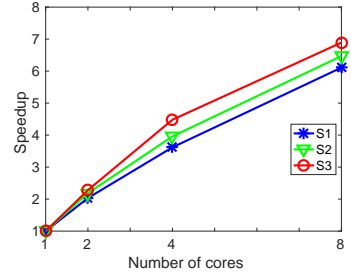}
  \caption{The speedup of \ppttf{} w.r.t the number of cores on three datasets.}\label{figure_speedup}
\end{figure}

\paragraph{Scalability}
Another metric used to measure a parallel algorithm is the scalability. 
To study the scalability of \ppttf, we test our model on three datasets by varying the number of cores from 1 to 8.   Figure \ref{figure_RMSE_Time} shows the test RMSE versus the running time for \ppttf{} model with different number of  cores (or sub-tensors) $P$.
The result demonstrates that the running time is approximately reduced to a half when the number of cores gets doubled. Note that the different curves of \ppttf{} eventually converge to the same solution.
To see more clearly, we compute the speedup relative to the running time with 1  core ($P=1$)  by varying the number of  cores from 1 to 8. Here, we set RMSE = 0.90 as a baseline. The results on S1-S3 are shown in Figure~\ref{figure_speedup}. We can intuitively see that \ppttf{} achieves nearly linear speedup and the speedup ratio increases with the number of ratings. The increased speedup from S1 to S3 is probably caused by the increase of data density.

\section{Conclusion}
In this paper, we present \ppttf{} by deriving a stochastic ADMM algorithm to calculate the latent factors of probabilistic temporal tensors. We propose a new data split strategy to divide the large-scale problem into several independent sub-problems along the user dimension. Then we use the parallel ADMM framework to decompose these sub-tensors in parallel. Experiments on real world data sets demonstrate that our \ppttf{} model outperforms the traditional CP decomposition, PTTF and PPMF model in terms of efficiency and scalability.

\bibliographystyle{IEEEtran}

\bibliography{prob_tensor}

\section*{Appendix}
We show the proof to Theorem 1 in the following.
  \subsection{Proof of Theorem 1 }\label{appendix_proof_H_L}
  \begin{proof}
  The constructed function $h(\cdot)$ in (13) 
  can be written as
   \begin{equation*}
  \begin{split}
&h(\bm{A}^p,\bm{B}^p,\bm{C}^p,\tau_t|\bm{A}^p_t,\bm{B}^p_t,\bm{C}^p_t)\\
&=\sum_{(i,j,k)\in\Omega^p}\hat{h}_{i,j,k}({A}_i^p,{B}_j^p,{C}_k^p, \tau_t|\bm{A}^p_t,\bm{B}^p_t,\bm{C}^p_t),
\end{split}
\end{equation*}
where
  \begin{equation*}
  \begin{split}
\hat{h}_{i,j,k}&({A}_i^p,{B}_j^p,{C}_k^p, \tau_t|\bm{A}^p_t,\bm{B}^p_t,\bm{C}^p_t)\\
=&\hat{f}_{i,j,k}(({A}_i^p)_t,({B}_j^p)_t,({C}_k^p)_t)\\
+&\nabla_{{A}_i^p}\hat{f}_{i,j,k}(({A}_i^p)_t,({B}_j^p)_t,({C}_k^p)_t) ({A}_i^p-({A}_i^p)_t)^T\\
+&\nabla_{{B}_j^p}\hat{f}_{i,j,k}(({A}_i^p)_t,({B}_j^p)_t,({C}_k^p)_t) ({B}_j^p-({B}_j^p)_t)^T\\
+&\nabla_{{C}_k^p}\hat{f}_{i,j,k}(({A}_i^p)_t,({B}_j^p)_t,({C}_k^p)_t) ({C}_k^p-({C}_k^p)_t)^T\\
+&[1/(2m_i\tau_t)]\|{A}_i^p-({A}_i^p)_t\|_2^2\\
+&[1/(2n_j\tau_t)] \|{B}_j^p-({B}_j^p)_t\|_2^2\\
 +&[1/(2z_k\tau_t)]\|{C}_k^p-({C}_k^p)_t\|_2^2.
\end{split}
\end{equation*}
Here, $m_i$ denotes the number of ratings related to $A_i^p$ in $\bm{\mathcal X}^p$, $n_j$ and $z_k$ are similarly defined.

Then, we have
\begin{equation*}
\begin{split}
L^p&(\bm{A}^p,\bm{B}^p,\bm{C}^p,(C_0^p)_t,(\bm{\Theta}_B^p)_t, (\bm{\Theta}_C^p)_t, \bm{\overline{B}}_t, \bm{\overline{C}}_t)\\
&-H^p(\bm{\mathcal M}^p,(C_0^p)_t, (\bm{\Theta}_B^p)_t, (\bm{\Theta}_C^p)_t ,\bm{ \overline{B}}_t, \bm{\overline{C}}_t, \tau_t|\bm{\mathcal M}^p_t) \\
= &f\left(\bm{A}^p,\bm{B}^p,\bm{C}^p\right)-h(\bm{A}^p,\bm{B}^p,\bm{C}^p, \tau_t|\bm{A}^p_t,\bm{B}^p_t,\bm{C}^p_t)\\
=&\sum_{(i,j,k)\in\Omega^p}[\hat{f}_{i,j,k}({A}_i^p,{B}_j^p,{C}_k^p)\\
&-\hat{h}_{i,j,k}({A}_i^p,{B}_j^p,{C}_k^p, \tau_t|\bm{A}^p_t,\bm{B}^p_t, \bm{C}^p_t)].
\end{split}
\end{equation*}

For clarity, we denote ${A}_i^p,{B}_j^p,{C}_k^p,({A}_i^p)_t,({B}_j^p)_t$ and $({C}_k^p)_t$ as $\bm a, \bm b,\bm c,\bm a_t,\bm b_t$ and $\bm c_t$, respectively. Then we have
\begin{equation*}
\begin{split}
\hat{f}_{i,j,k}(\bm a,\bm b ,\bm c)=&\frac{1}{2}(x_{ijk}^p-<\bm a-\bm a_t+\bm a_t,\\
&\bm b-\bm b_t+\bm b_t ,\bm c-\bm c_t+\bm c_t>)^2\\
=&\hat{f}_{i,j,k}(\bm a_t,\bm b_t ,\bm c_t)\\
+&\nabla_{\bm a}\hat{f}_{i,j,k}(\bm a_t,\bm b_t,\bm c_t) (\bm a-\bm a_t)^T\\
+&\nabla_{\bm b}\hat{f}_{i,j,k}(\bm a_t,\bm b_t,\bm c_t) (\bm b-\bm b_t)^T\\
+&\nabla_{\bm c}\hat{f}_{i,j,k}(\bm a_t,\bm b_t,\bm c_t) (\bm c-\bm c_t)^T\\
+&\bm o(\bm a,\bm b ,\bm c),
\end{split}
\end{equation*}
where $\bm o(\bm a,\bm b ,\bm c)$ contains all the second to sixth order terms.

We have the following properties by mainly using Cauchy inequality and the hypothesis $\|\bm a-\bm a_t\|_2^2\leq \delta^2,\|\bm b-\bm b_t\|_2^2\leq \delta^2, \|\bm c-\bm c_t\|_2^2\leq \delta^2$:
\begin{equation*}
\begin{split}
-(&x_{ijk}^p-<\bm a_t,\bm b_t,\bm c_t>)<\bm a-\bm a_t,\bm b-\bm b_t,\bm c_t>\\
&\leq \bm|\epsilon_{ijk}\bm |(\bm a-\bm a_t)((\bm b-\bm b_t)\ast\bm c_t)^T\\
&\leq\frac{1}{2}\bm|\epsilon_{ijk}\bm |(\|\bm a-\bm a_t\|_2^2+ \|\bm b-\bm b_t\|^2_2\|\bm c_t\|^2_2);
\end{split}
\end{equation*}
\begin{equation*}
\begin{split}
-(&x_{ijk}^p-<\bm a_t,\bm b_t,\bm c_t>)<\bm a-\bm a_t,\bm b-\bm b_t,\bm c-\bm c_t>\\
&\leq \bm|\epsilon_{ijk}\bm |(\bm a-\bm a_t)((\bm b-\bm b_t)\ast(\bm c-\bm c_t))^T\\
&\leq\frac{1}{2}\bm|\epsilon_{ijk}\bm |(\|\bm a-\bm a_t\|_2^2+ \|\bm b-\bm b_t\|^2_2\|\bm c-\bm c_t\|^2_2)\\
&\leq\frac{1}{2}\bm|\epsilon_{ijk}\bm |(\|\bm a-\bm a_t\|_2^2+ \|\bm b-\bm b_t\|^2_2\delta^2);
\end{split}
\end{equation*}
where $\epsilon_{ijk}=x_{ijk}^p-<\bm a_t,\bm b_t,\bm c_t>$.
\begin{equation*}
\begin{split}
(a_1+a_2+\ldots+a_n)^2\leq n(a_1^2+a_2^2+\ldots+a_n^2);
\end{split}
\end{equation*}
\begin{equation*}
\begin{split}
(<&\bm a-\bm a_t,\bm b-\bm b_t,\bm c-\bm c_t>)^2\\
&\leq\frac{1}{4}(\|\bm a-\bm a_t\|_2^2+ \|\bm b-\bm b_t\|^2_2\delta^2)^2\\
&\leq\frac{1}{4}(\delta^2+ \delta^4)(\|\bm a-\bm a_t\|_2^2+ \|\bm b-\bm b_t\|^2_2\delta^2);
\end{split}
\end{equation*}
\begin{equation*}
\begin{split}
&(<\bm a-\bm a_t,\bm b-\bm b_t,\bm c_t>)^2\\
&\leq\frac{1}{4}(\|\bm a-\bm a_t\|_2^2+ \|\bm b-\bm b_t\|^2_2\|\bm c_t\|^2_2)^2\\
&\leq\frac{1}{4}(\delta^2+ \delta^2\|\bm c_t\|^2_2)(\|\bm a-\bm a_t\|_2^2+ \|\bm b-\bm b_t\|^2_2\|\bm c_t\|^2_2);
\end{split}
\end{equation*}
\begin{equation*}
\begin{split}
(<\bm a-\bm a_t,\bm b_t,\bm c_t>)^2&\leq(\|\bm a-\bm a_t\|_2\|\bm b_t\ast\bm c_t\|_2)^2\\
&=\|\bm b_t\ast\bm c_t\|_2^2\|\bm a-\bm a_t\|_2^2.
\end{split}
\end{equation*}
Using the above six properties, we can prove that
\begin{equation*}
\begin{split}
\bm o(\bm a,\bm b ,\bm c)\leq&\pi_A\|\bm a-\bm a_t\|_2^2+\pi_B\|\bm b-\bm b_t\|_2^2\\
&+\pi_C\|\bm c-\bm c_t\|_2^2,
\end{split}
\end{equation*}
where $\pi_A,\pi_B,\pi_C$ are constants which depend on $\bm a_t$, $\bm b_t$, $\bm c_t$ and $\delta^2$.

If we let
\begin{equation*}
\begin{split}
\frac{1}{\tau_t}\geq \max\{2m_i\pi_A, 2n_j\pi_B, 2z_k\pi_C\},
\end{split}
\end{equation*}
then we can prove that
\begin{equation*}
\begin{split}
H^p&(\bm{\mathcal M}^p,(C_0^p)_t, (\bm{\Theta}_B^p)_t, (\bm{\Theta}_C^p)_t ,\bm{ \overline{B}}_t, \bm{\overline{C}}_t, \tau_t|\bm{\mathcal M}^p_t) \\
&\geq L^p(\bm{A}^p,\bm{B}^p,\bm{C}^p,(C_0^p)_t,(\bm{\Theta}_B^p)_t, (\bm{\Theta}_C^p)_t, \bm{\overline{B}}_t, \bm{\overline{C}}_t).
\end{split}
\end{equation*}
The second property in Theorem 1
can be easily proved.
  \end{proof}
\end{document}